\documentclass{article}
\usepackage{spconf,amsmath,graphicx,hyperref}
\usepackage{pifont}  
\newcommand{\cmark}{\ding{51}}  % 对勾 ✓
\newcommand{\xmark}{\ding{55}}  % 叉叉 ✗
\usepackage{amsmath}
\usepackage{amssymb}
\usepackage{amsfonts}
\usepackage{booktabs}   % 提供 \toprule,\midrule,\bottomrule,\cmidrule
\usepackage{makecell}   % 提供 \makecell，对单元格内换行很方便
\usepackage{stfloats}
\usepackage{marvosym}   % 提供 \Letter

\title{HiCT: High-Precision 3D CBCT Reconstruction from a Single X-ray}
\address{ }
\name{Wen Ma\textsuperscript{1}, 
      Jiaxiang Liu\textsuperscript{1}, 
      Zikai Xiao\textsuperscript{1}, 
      Ziyang Wang\textsuperscript{1},
      Feng Yang\textsuperscript{2},
      Zuozhu Liu\textsuperscript{1\Letter}
      }
\vspace{-2cm} 
\address{\textsuperscript{1} Zhejiang University, 
         \textsuperscript{2} Angelalign Technology Inc.}
\begin{document}
%\ninept
%
\maketitle
\begin{abstract}
Accurate 3D dental imaging is vital for diagnosis and treatment planning, yet CBCT’s high radiation dose and cost limit its accessibility. Reconstructing 3D volumes from a single low-dose panoramic X-ray is a promising alternative but remains challenging due to geometric inconsistencies and limited accuracy. We propose HiCT, a two-stage framework that first generates geometrically consistent multi-view projections from a single panoramic image using a video diffusion model, and then reconstructs high-fidelity CBCT from the projections using a ray-based dynamic attention network and an X-ray sampling strategy. To support this, we built XCT, a large-scale dataset combining public CBCT data with 500 paired PX-CBCT cases. Extensive experiments show that HiCT achieves state-of-the-art performance, delivering accurate and geometrically consistent reconstructions for clinical use. Code will be publicly released.
% Accurate 3D dental imaging is crucial for diagnosis and treatment planning. While high-resolution Cone-Beam CT (CBCT) provides essential volumetric detail, its significant radiation exposure and cost limit its widespread use. Reconstructing high-fidelity 3D volumes from a single, low-dose panoramic X-ray (PX) offers a promising and accessible solution. However, this 2D-to-3D conversion is challenging; existing methods frequently exhibit geometric inconsistencies and lack the reconstruction accuracy required for clinical applications.
% To bridge the gap, we introduce \textbf{HiCT}, the framework that reconstructs a high-quality 3D CBCT from a 2D panoramic X-ray.
% The framework consists of two stages. 
% In the first stage, HiCT synthesizes a set of geometrically consistent multi-view projections by leveraging a video diffusion model with implicit 3D priors and explicit camera position. 
% In the second stage, HiCT reconstructs a high-precision 3D CBCT from the multi-view projections using a reconstruction framework guided by Ray-based Dynamic Attention (RDA) and X-ray Sampling strategy.
% To demonstrate the effectiveness of HiCT, we constructed a large-scale XCT dataset by integrating publicly available CBCT data with 500 high-quality PX-CBCT paired cases acquired through in-house collection. Extensive experiments demonstrate that HiCT achieves state-of-the-art performance, producing high precision reconstructions and strong geometric consistency.

\end{abstract}
\begin{keywords}
3D, Reconstruction, X-Ray, Neural Rendering, Multi-view Synthesis
\end{keywords}
\section{Introduction}
\label{sec:intro}

Cone-beam CT (CBCT) provides unparalleled 3D detail for anatomical analysis and treatment planning, making it a key tool in dental diagnostics \cite{liu2023toothsegnet}. However, its high radiation dose—about ten times that of panoramic X-ray (PX)—and costly equipment limit its widespread use \cite{px}. While PX offers a low-dose, affordable, and easily accessible alternative, its 2D nature and arc-shaped projection distortions hinder accurate assessment of complex maxillofacial structures \cite{Meshsnet}. Researchers have explored reconstructing diagnostic-quality 3D volumes from a single PX image, aiming to reduce radiation exposure while enabling accurate and affordable 3D imaging in resource-limited settings \cite{mw,liu2023deep}.

%·、1 projection 2 input 区别 全景片的好处 更多信息 获取成本较低  3 上面别人的 下面我们的 4 因果关系体现一下 5

% 问题 1:多视图几何一致性差+牙体解剖结构特异性不足  2:像素级射线采样效率低 3:对 X 射线成像物理机理适应性差
Nevertheless, reconstructing 3D CBCT from a single PX faces three key challenges that hinder clinical reliability:
\textbf{(i) Multi-view inconsistency. }Existing diffusion-based view synthesis often ignores 3D priors, making it difficult to maintain consistency in complex dental structures (e.g., enamel, roots, jawbone).
\textbf{(ii) Inefficient sampling. } Efficiently acquiring sufficient viewpoint information is a key factor in ensuring the accuracy of 3D reconstruction. NeRF-based methods use random ray sampling that wastes computation in low-density regions and undersamples critical anatomical areas, leading to poor detail reconstruction.
\textbf{(iii) Inadequate X-ray modeling.} Many methods inherit surface-reflectance models from RGB imaging, overlooking X-ray physics of penetration and attenuation, thus failing to recover internal anatomical structures essential for accurate CBCT.

To address the above challenges, we propose HiCT, a novel two-stage framework for reconstructing high-quality 3D CBCT from a single PX. 
In Stage 1, HiCT synthesizes a set of geometrically coherent multi-view projections to overcome the inconsistencies of traditional methods. Instead of generating views independently, we employ a video diffusion model that treats the projections as a continuous sequence, inherently learning the 3D priors needed for inter-view consistency. 
In Stage 2, HiCT reconstructs the CBCT volume from the synthesized multi-view projections using two key components. First, to focus on key reconstruction areas, we introduce X-ray Sampling. A density-driven hybrid strategy that builds a 3D occupancy grid along each ray, skipping zero-density regions and focusing samples in high-absorption areas. Second, to capture the structural relationships, we propose the Ray-based Dynamic Attention (RDA) network, which models sampled points along each ray as a sequence and applies multi-head self-attention to capture structural dependencies. 
% In Stage 2, HiCT reconstructs a high-quality CBCT from the multi-view X-ray projections generated in Stage 1.
% First, to overcome the inefficiency of standard sampling, we propose a density-driven hybrid sampling strategy tailored to X-ray physics (\textbf{X-ray Sampling}). Unlike conventional random sampling, we generate a 3D occupancy grid along each X-ray to skip zero-density regions and employ stratified sampling to construct an absorption distribution with a derived probability density function. It adaptively skips low-density regions (e.g., air, soft tissue) and concentrates samples in high-absorption areas (e.g., bone), enhancing reconstruction accuracy with reduced cost. 
% Second, to model the physical properties of X-ray attenuation more accurately, we introduce the Ray-based Dynamic Attention (\textbf{RDA}) network.
% RDA replaces the simple MLP used in prior work, which fails to capture the structural dependencies between points along a ray. RDA treats the sampled points on each ray as a sequence and applies a multi-head self-attention mechanism to model their interrelationships. RDA allows the network to adaptively weigh the most informative points, achieving a more fine-grained and physically plausible reconstruction of the CBCT volume.
\begin{figure*}[b]
\centering

\includegraphics[width=1.0\linewidth]{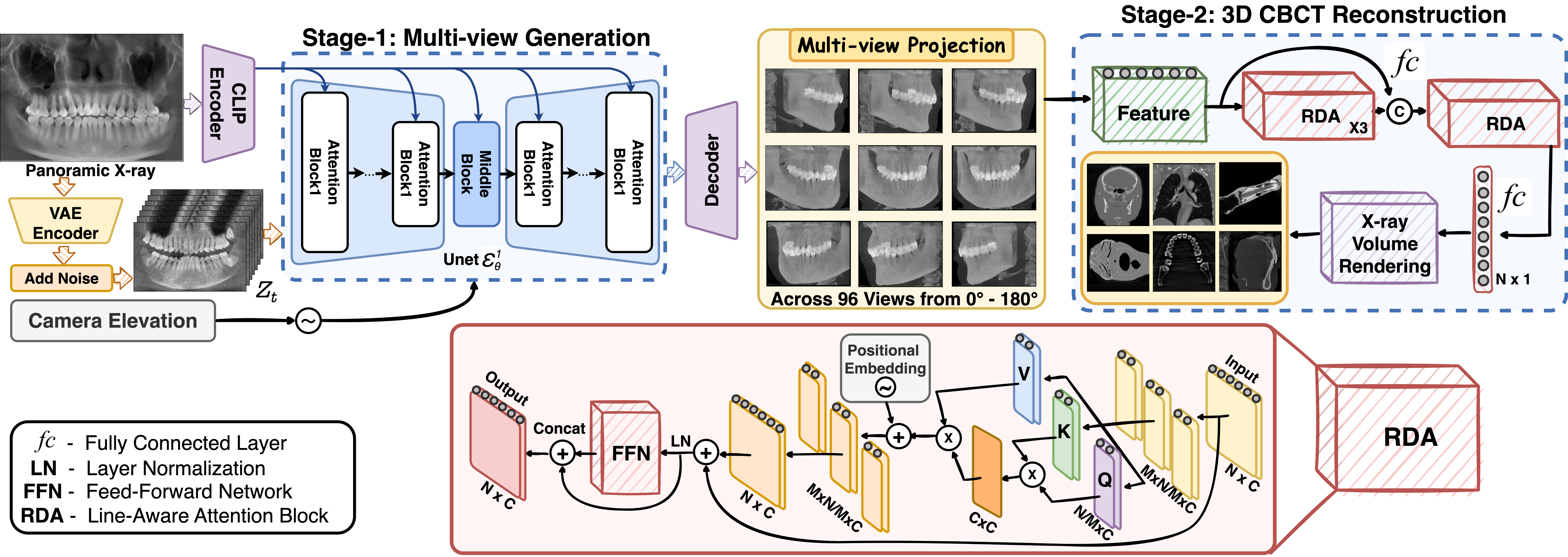} 
\caption{The pipeline of HiCT.}
\label{fig:pipeline}
\vspace{-1em}
\end{figure*}

We validate HiCT through extensive qualitative and quantitative experiments. To support robust evaluation, we construct a large-scale XCT dataset with 500 CBCT–PX pairs, an order of magnitude larger than existing datasets, along with integrated public CBCT sources. Experimental results show that HiCT consistently outperforms state-of-the-art methods in single-view CBCT reconstruction. Our main contributions are as follows:
\begin{itemize}
    \item HiCT employs a video diffusion model with camera pose conditioning to generate geometrically consistent multi-view projections, effectively enforcing 3D priors and inter-view coherence. 

    \item HiCT reconstructs the CBCT volume using two key modules: a density-driven X-ray Sampling strategy that focuses on high-absorption regions while skipping low-density areas, and the Ray-based Dynamic Attention (RDA) network that captures structural dependencies along rays for fine-grained reconstruction.

    \item We construct \textbf{XCT}, a large-scale dataset that integrates multiple public CBCT datasets with 500 in-house PX-CBCT pairs, serving as a valuable benchmark for future research in 2D-to-3D dental imaging.
\end{itemize}

\section{Related Work}
\label{sec:format}

% 近年来+分为两类 单个图像视角合成 + 3D CBCT 重建
\textbf{Novel view synthesis (NVS):} NVS aims to infer unobserved image regions from single or sparse viewpoints. Early methods in the natural image domain include CNN‐based single‐image view generation, which learn disentangled latent spaces to project and edit images for novel views. Subsequent work leveraged GANs to recover occluded content, but these methods rely on sufficient pixel correspondences between views, an assumption that fails in X‐ray imaging due to extremely sparse overlap~\cite{yu2021pixelnerf}. Later, generative approaches improved hallucination and consistency: Li et al.~\cite{li2022infinitenature} enhanced invisible region synthesis with GANs, while transformer-based models~\cite{kulhanek2022viewformer} improved multi-view coherence. Recently, diffusion models have advanced the field: Zero1to3~\cite{zero123} enables zero-shot generalization, GeNVS~\cite{GeNVS} and SyncDreamer~\cite{Syncdreamer} embed 3D features, Tseng et al.~\cite{tseng} enforce view consistency via attention, and Wonder3D~\cite{long2024wonder3d} and SV3D~\cite{voleti2024sv3d} exploit cross-domain and video diffusion for stronger spatial–temporal consistency.

\textbf{CBCT reconstruction:} Methods include analytical and optimization-based approaches. Analytical methods invert the Radon transform and work well with dense projections but fail in sparse views. Optimization-based methods adopt MAP estimation with handcrafted priors, achieving higher accuracy but at heavy computational cost. Recent deep learning methods, including CNNs~\cite{lin2023learning} and diffusion models~\cite{chung2023solving}, improve quality yet require large paired datasets. NeRF~\cite{nerf} has advanced 3D scene reconstruction, inspiring sparse-view variants: DS-NeRF~\cite{ds-nerf} enhances quality with depth supervision but still needs many views; Zerorf~\cite{shi2024zerorf} incorporates depth priors yet struggles to generalize; and NeuSurf~\cite{huang2024neusurf} exploits signed distance functions but degrades with limited data. To address these challenges, we propose a NeRF-based method for efficient, high-fidelity CBCT reconstruction.

% \begin{figure*}[t]
% \centering
% \includegraphics[width=1\linewidth]{Figures/PipeLine-all.png} % Reduce the figure size so that it is slightly narrower than the column.
% \caption{
% Overview of HiCT. Stage-1 generates multi-view projections; Stage-2 reconstructs 3D CBCT from them.

% Overview of the HiCT framework for reconstructing high-resolution 3D CBCT from a single PX. Stage-1 generates a 180° multi-view sequence with 3D awareness by introducing temporal consistency from orbital video generation to ensure geometric coherence. Stage-2 adopts an adaptive hybrid sampling strategy based on X-ray physical properties for accurate and efficient CBCT reconstruction.
% }
% \label{fig:pipeline}
% \end{figure*}

\section{Method}
\label{sec:pagestyle}
\textbf{Stage-1: View Synthesis From a Single PX.}
To enhance 3D perception, we innovatively adopted the continuous frame training strategy of video diffusion models. Given the input panorama \(I_i\), we first project it into the latent space using the VAE encoder of the video diffusion model, and channel-wise concatenate it with the noisy latent sequence, which encourages synthesized multi-view images to preserve the identity and intricate details of the input image. In addition, we incorporate the CLIP embeddings of the input condition image into the diffusion model through a cross-attention mechanism. Within each transformer block, the CLIP embedding matrix acts as both the key and the value for the cross-attention layers, while the block feature map serves as the query. In this way, the high-level semantic information of the input image is propagated into the video diffusion model. To mitigate view inconsistency and sampling artifacts, we explicitly inject camera elevation\(e\) into the conditioning pathway.

% To resolve the multi-view inconsistency prevalent in existing methods, which neglect 3D prior knowledge and independently model inter-view relationships, leading to unreliable reconstructions. We redefine the single image to new views generation task as a 3D-aware sequential multi-view projection generation problem, and incorporate camera pose conditioning into HiCT to produce high-resolution, geometrically consistent projection sequences.

In HiCT, the denoiser is a 3D U-Net $\epsilon_{\theta}^{1}(z_t; I, t, e)$ that predicts the Gaussian noise added to latent sequences. 
Given a multi-view image sequence $J$, a pretrained VAE encoder $\mathcal{E}$ maps each frame to latents $\mathbf{z}$. 
At step $t$, the forward diffusion adds $\epsilon \sim \mathcal{N}(0,\mathbf{I})$ to obtain a noisy latent $z_t$. 
The network estimates $\epsilon$ conditioned on the input image $I$ and elevation $e$, optimized with a weighted MSE:
\vspace{-0.5em}
\begin{equation}
\mathcal{L}_{\mathrm{Stage\mbox{-}1}}
= \mathbb{E}_{I,J,e,t,\epsilon}
\!\left[\left\|\,w(t)\big(\epsilon_{\theta}^{1}(z_t; I, t, e) - \epsilon\big)\right\|_{2}^{2}\right],
\label{eq:stage1_loss}
\end{equation}
\vspace{-0.5em}
where $w(t)$ is the timestep weight.
\vspace{0.5em}

% In general, the denoising neural network in the HiCT model can be represented as $\epsilon_{\theta}^{1}(z_t; I, t, e)$.
% Given the multi-view image sequence \(J\), the pre-trained VAE encoder \(\mathcal{E}(\cdot)\) first extracts the latent code of each image to constitute a latent code sequence \(\mathbf{z}\). Next, Gaussian noise \(\epsilon \sim \mathcal{N}(0, I)\) is added to \(\mathbf{z}\) through a typical forward diffusion procedure at each time step \(t\) to obtain the noisy latent code \(z_t\). The 3D UNet \(\epsilon_{\theta}^{1}(z_t; I, t, e)\) with parameter \(\theta\) is trained to estimate the added noise \(\epsilon\) based on noisy latent code \(z_t\). 

% input image condition \(I\) and elevation angle \(e\) through the standard mean square error (MSE) loss:
% \begin{equation}
% \mathcal{L}_{\mathrm{Stage\mbox{-}1}}
% = \mathbb{E}_{I,J,e,t,\epsilon}
% \Bigl[\bigl\|\,w(t)\bigl(\epsilon_{\theta}^{1}(z_t; I, e, t) - \epsilon\bigr)\bigr\|_{2}^{2}\Bigr],
% \label{eq:stage1_loss}
% \end{equation}
% where \(w(t)\) is the corresponding weighting factor.

\textbf{Stage-2: 3D CBCT Reconstruction}
Unlike existing methods, we design a density-driven hybrid sampling strategy tailored to X-ray’s sparse–dense distribution. As shown in Fig.~\ref{fig:pipeline}, we first build a 3D occupancy grid and skip near-vacuum intervals ($w\!\approx\!0$). In the remaining segments, $N_1$ coarse samples $\{t_i\}$ are drawn via stratified sampling, and their densities $\{w_i\}$ form a normalized PDF. Guided by this PDF, we resample $N_2$ fine points within each segment using systematic resampling:
\vspace{-0.5em}
\begin{equation}
    t'_{i,k} = z_i + \Bigl(\bigl(v+\tfrac{k}{N_2}\bigr)-\lfloor v+\tfrac{k}{N_2}\rfloor\Bigr)\,(z_{i+1}-z_i).
\end{equation}
\vspace{-0.3em}
The combined $N_1+N_2$ samples are then arranged in a circular scanning layout and passed to the renderer. This approach skips redundant empty regions and adaptively densifies high-absorption zones, thereby improving detail fidelity and noise robustness in CBCT reconstruction.

\begin{table*}[hbtp]
\centering
\resizebox{\linewidth}{!}{
\begin{tabular}{l *{5}{cc} | cc}
   \specialrule{1pt}{0pt}{2pt}   
    % 第一行：左上角竖表头 + 9 个方法各合并两列
    \makecell[l]{Method}
      & \multicolumn{2}{c}{FDK }
      & \multicolumn{2}{c}{ASD }
      & \multicolumn{2}{c}{SART}

      & \multicolumn{2}{c}{NeRF  }
      & \multicolumn{2}{c|}{NAF }
     & \multicolumn{2}{c}{\textbf{HiCT}} \\
    % 在每个合并列下画分隔线
    \cmidrule(lr){2-3}  \cmidrule(lr){4-5}  \cmidrule(lr){6-7}
    \cmidrule(lr){8-9}  \cmidrule(lr){10-11}\cmidrule(lr){12-13}

    % 第二行：各列小标题
      Scene & PSNR & SSIM  
      & PSNR & SSIM  
      & PSNR & SSIM  
      & PSNR & SSIM  
      & PSNR & SSIM  
      & PSNR & SSIM  \\
    \specialrule{1pt}{0pt}{2pt}
    % —— 在这里开始填你的数据 —— 
    Head & 25.16 & 0.7154    & 35.26  & 0.8106  & 34.87 & 0.8056  & 34.81 & 0.8618 
    &\underline{35.32} &\underline{0.9014} & \textbf{39.69}  & \textbf{0.9880} \\
    
    Chest & 22.89 & 0.6861   & 31.12 & 0.8421   & 32.17 &  0.8584 &\underline{33.27} & 0.8629  
    &  32.30  & \underline{0.9158} & \textbf{37.87}  & \textbf{0.9719}  \\
    
    Leg  & 24.47 &  0.6690   & 35.39 & 0.8213   & 35.29 & 0.8409  & 32.01 & 0.8557  
    &   \underline{36.57}  & \underline{0.9401}   & \textbf{43.95}   &  \textbf{0.9978}  \\
    
    Carp & 22.31 & 0.7273    & 33.06 & 0.8337   & 36.74 & 0.8282  & 32.29 & 0.8490 
    & \underline{38.16} & \underline{0.9029}   &\textbf{46.72}  &\textbf{0.9902} \\

    Foot & 23.02 & 0.7815   & 29.89 & 0.8209  & 30.29 & 0.8296 
    & \underline{33.03} & \underline{0.9312} &31.63 & 0.9101 &\textbf{37.98}&\textbf{0.9403} \\
    
    Pelvis & 23.84 & 0.5221   & 34.26 & \underline{0.9104}  & \underline{34.38} & 0.8481  & 31.72 & 0.8970 & 32.01& 0.9045 &   \textbf{41.40} &\textbf{0.9870} \\
    
    Pancreas & 22.63 & 0.5233 & 18.30 & 0.7201  & 18.36 & 0.7001  & 17.73 & 0.8614 
    &\underline{23.41} & \underline{0.8891} &  \textbf{25.98}  &\textbf{0.9598} \\
    
    Abdomen & 22.91 & 0.6030  & 31.46 & 0.8220  & 31.40 & 0.8017  & 29.71 & \underline{0.9149}
    & \underline{34.05}& 0.8905 & \textbf{39.01}  &\textbf{0.9688} \\

    Jaw & 26.57 & 0.7815   & 33.25 & 0.8066   & 33.12 & 0.8300  & 33.05 & 0.8912 & \underline{33.89} &\underline{0.9316}&  \textbf{38.91}  &\textbf{0.9524} \\
    
    Tooth (Ours) & 21.42 & 0.5221 & 31.27 &  0.8501 & 31.55 & 0.8466 & 31.82 & 0.8701 
    &\underline{36.71} & \underline{0.9295} &\textbf{40.55}  &\textbf{0.9783} \\
    
    \specialrule{1pt}{0pt}{2pt}
    \textbf{Average} & 23.52 & 0.6131 & 31.73 & 0.8238 & 31.81  & 0.8189 & 30.94  &   0.8795 
    & 33.41 & 0.9115 &\textbf{39.21}  &\textbf{0.9735} \\
    
    \specialrule{1pt}{0pt}{0pt}
\end{tabular}}
\caption{Comparisons on the 3D CBCT reconstruction task. The best and second-best results are marked in bold and underlined.}
\label{tab2}
\vspace{-1em}
\end{table*}

\textbf{Ray-Based Dynamic Attention Framework.}
X-ray attenuation reveals internal structures critical for CBCT reconstruction, yet prior NeRF-based methods rely on MLPs that treat ray samples independently, ignoring their sequential dependencies~\cite{aenerf,Nerf-det++}. To address this, we propose a RDA framework. Sampling coordinates $P$ are first encoded via a hash encoder $\mathcal{H}$ into features $F=\mathcal{H}(P)$, which are processed with residual connections. Each RDA block applies nonlinear fusion followed by a line-segment multi-head self-attention module, where samples along a ray are treated as a sequence and dependencies are adaptively weighted. A mapping head then predicts radiodensity $D$ for each point, enabling fine-grained, coherent modeling of the X-ray transmission process. By capturing the interactions among sampled points within each ray segment, the RDA not only dependency–awarely perceives the continuous correlations between points, but also achieves global information fusion through weighted multi-head self-attention. As a result, the network exhibits heightened sensitivity to attenuation differences and subtle structural variations along different X-ray paths, enabling more precise and coherent modeling of the implicit neural radiodensity field.

\begin{figure*}[t]
\centering
\vspace{-1em}
\includegraphics[width=1.0\linewidth]{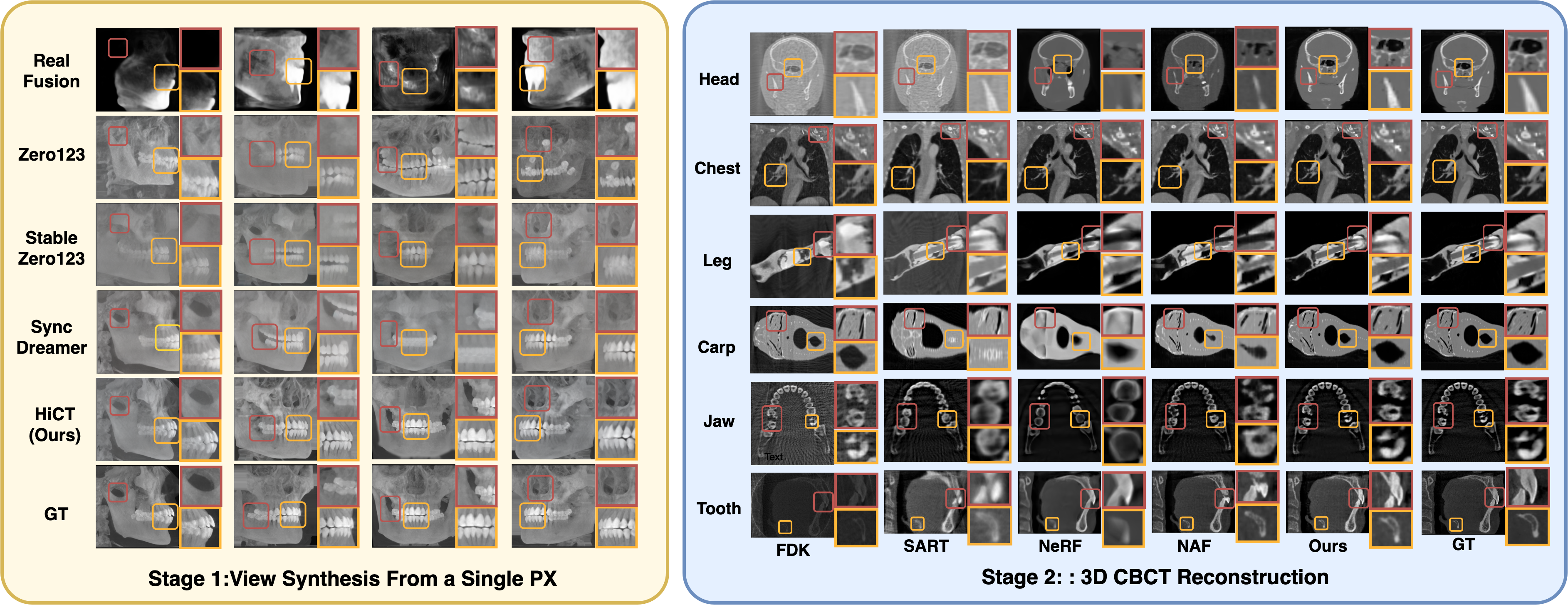} % Reduce the figure size so that it is slightly narrower than the column.
\vspace{-2em}
\caption{HiCT demonstrates excellent visual consistency and boundary clarity in the results of multi - view projection generation stage and 3D CBCT generation stage when compared with GT.}
\label{fig:visualize1}
\vspace{-0.8em}
\end{figure*}

% \begin{figure*}[hbtp]
% \centering
% \includegraphics[width=0.8\linewidth]{Figures/visualize2.png} % Reduce the figure size so that it is slightly narrower than the column.
% \caption{Visualization of the CBCT reconstruction. Please zoom in for more details.}
% \label{fig:visualize2}
% \end{figure*}

\vspace{-1em}
\section{Experiments}
\label{sec:typestyle}
\vspace{-1em}

\textbf{Dataset:} We evaluated the effectiveness of HiCT on the XCT dataset, which consists of two parts: in-house data comprising 500 high-quality PX-CBCT paired samples, an order of magnitude larger than existing public paired datasets. Public data integrating 162 CBCT data of the head, chest, lower limbs, and dentition sourced from LIDC-IDRI, COVIDx, TCIA, and The Open Scientific Visualization Dataset, respectively. 
% We first performed rigid registration based on bony landmarks and standardized both spatial resolution and intensity range. Then we used the TIGRE~\cite{tigre} to generate 0°–180° full‐angle projections with 3\% noise, thereby ensuring both clinical representativeness and experimental controllability. Given that PX is an imaging modality unique to dental practice, public CBCT data are used only in stage-2 experiments.

% \textbf{Metrics:}Following existing works~\cite{lin2023learning,NAF,24}, we evaluate our HiCT using five metrics. PSNR measures pixel-level reconstruction accuracy, while SSIM  assesses structural similarity. LPIPS, grounded in deep perceptual features, captures human-aligned visual differences. For geometric validation, IoU and Dice quantify shape and boundary consistency. 
% These metrics, validated in extensive clinical studies and recognized by international guidelines~\cite{dohmen2025similarity}, are widely regarded as reliable and clinically applicable. We compute these metrics for each generated slice, and report the mean values for comparison.

\textbf{Comparison Methods:}
To evaluate HiCT's performance comprehensively, we compared it with a range of advanced methods, aiming to assess its advantages across diverse technical frameworks. In stage-1, for multi-view projection generation we benchmarked RealFusion, Zero123, Zero123++, Zero123-XL, and Stable-Zero123. We also evaluated diffusion-based SyncDreamer as a view synthesis baseline. In stage-2, for the generation of 3D CBCT via traditional volumetric reconstruction, we adopt FDK, ASD, and SART; for neural field reconstruction, we tested NeRF and NAF.

\vspace{-1em}
\subsection{Main Results} 
\vspace{-0.5em}
\textbf{Quantitative Results:}
Table~\ref{tab2} shows that HiCT significantly outperforms all baseline methods in both PSNR and SSIM across all dataset scenes. On average, HiCT achieves 39.21 dB PSNR (5.8 dB higher than the next-best method, NAF at 33.41 dB) and 0.9735 SSIM (0.06 higher than NAF’s 0.9115). In specific regions like the "Leg," HiCT attains 43.95 dB PSNR and 0.9978 SSIM, surpassing the best baseline by over 7 dB. Even in challenging cases like the "Pancreas," HiCT (25.98 dB PSNR, 0.9598 SSIM) significantly outperforms ASD (18.30 dB, 0.7201). These results highlight HiCT’s superior capability in preserving both peak signal fidelity and structural similarity, demonstrating robust and high-quality 3D CBCT reconstruction. 
Table~\ref{tab1} compares HiCT with state-of-the-art methods across five key metrics. HiCT achieves the highest PSNR (21.014), significantly outperforming SyncDreamer (16.262), indicating better image quality and finer detail preservation. It also leads in SSIM (0.865), LPIPS (0.101) (minimal perceptual difference), and excels in segmentation metrics with IoU (0.816) and Dice (0.877), confirming its accuracy in multi-view synthesis.

\begin{table}[htbp]
  \centering
  \small            % 或 \footnotesize
\resizebox{\linewidth}{!}{
\begin{tabular}{l|ccccc}
    \specialrule{1pt}{0pt}{2pt}
    Method  &   PSNR  & SSIM  & LPIPS $\downarrow$   & IoU  & Dice \\
    \specialrule{1pt}{0pt}{2pt}
    RealFusion &  6.665 & 0.489   & 0.683 & 0.497 & 0.502   \\
    Zero123 & 10.574  & 0.571   & 0.505 & 0.539 & 0.601 \\
    Zero123++ & 11.925  & 0.668   & 0.442 & 0.640 & 0.623 \\
    Zero123-XL & 13.052  & 0.691   & 0.423 & 0.698 & 0.670 \\
    Stable-Zero123 & 15.129  & 0.701   & 0.396  & 0.733 & 0.737 \\
    SyncDreamer & 16.262  &  0.742  &  0.283 & 0.769 & 0.792 \\
    {\bfseries HiCT} & {\bfseries 21.014 } & {\bfseries 0.865 } &  {\bfseries 0.101 }
    &  {\bfseries 0.816  }&  {\bfseries 0.877 }\\
    \specialrule{1pt}{0pt}{2pt}
\end{tabular}}
\vspace{-1em}
\caption{Results of novel multi-view projections synthesis.}
% : Perceptual‐quality metrics (PSNR, SSIM, LPIPS) and geometric‐accuracy metrics (IoU, Dice).}
\label{tab1}
\vspace{-1em}
\end{table}

\textbf{Qualitative Results:}
In stage-1, we compared HiCT with six state-of-the-art algorithms (Fig.\ref{fig:visualize1}). RealFusion exhibits missing high-frequency textures and distortions. Zero123 generates coherent structures but lacks detail. Zero123++ improves with multi-scale fusion yet struggles with complex structures. Zero123-XL shows minor gains but retains ghosting artifacts. Stable-Zero123 improves sampling but suffers from ghosting and boundary issues. SyncDreamer enhances textures but lacks geometric consistency. In contrast, HiCT preserves complete geometry with crisp textures and fine details, maintaining semantic accuracy and view consistency.
In Stage-2, CBCT reconstruction methods show varied performance across datasets (Fig.\ref{fig:visualize1}). FDK has obvious artifacts. ASD/SART improve smoothness but lose texture/detail, distorting boundaries. NeRF restores overall morphology well but lacks fine-detail clarity. NAF reduces artifacts and improves consistency, yet boundary sharpness/details are limited. But HiCT yields sharp contours across regions, faithful fine-feature restoration, and excellent multi-view consistency.

\vspace{-1em}
\subsection{Ablation Study} 
\vspace{-0.5em}

The ablation results in Table~\ref{tab3} validate the contributions of X-ray sampling and RDA to CBCT reconstruction. The baseline achieves a PSNR of 34.40dB. Introducing X-ray sampling alone improves PSNR to 36.13dB, indicating its effectiveness in enhancing projection quality through optimized sampling. Adding RDA alone boosts PSNR to 37.69dB, with improvements particularly pronounced in regions requiring fine structural detail (e.g., enamel-dentin junctions). Combining both yields the best performance (39.21dB), a 4.81dB gain over the baseline, demonstrating their complementary benefits in improving reconstruction fidelity and consistency.

\begin{table}[h!]
\vspace{-0.5em}
\begin{center}
\begin{tabular}{cc|ccc}
\specialrule{1pt}{0pt}{2pt}
X-ray Sampling & RDA &  PSNR  ( $\uparrow$ ) & SSIM ( $\uparrow$ )  \\
\specialrule{1pt}{0pt}{2pt}
\xmark & \xmark & 34.40  & 0.9223  \\
\cmark & \xmark & 36.13  & 0.9437 \\
\xmark & \cmark & 37.69  & 0.9502  \\
\cmark & \cmark & {\bfseries 39.21} & {\bfseries 0.9735 } \\
\specialrule{1pt}{0pt}{2pt}
\end{tabular}
\end{center}
\vspace{-2em}
\caption{Ablation results respectively show the impact of density-driven X-ray mixed-sampling method and RDA on CBCT reconstruction quality.}
\label{tab3}
\vspace{-2.5em}
\end{table}

\section{Conclusions}
\label{sec:print}
\vspace{-1em}
In this work, we propose HiCT, the pioneering framework for reconstructing high-fidelity, geometrically consistent 3D CBCT from a single PX. The task is decomposed into two stages: Stage-1 leverages multi-frame modeling and explicit camera pose conditioning to generate multi-view projections, while stage-2 reconstructs CBCT volumes via X-ray–based hybrid sampling and ray attenuation simulation. To validate effectiveness, we evaluate HiCT on multiple public datasets and introduce a XCT dataset with 500 paired PX-CBCT cases. Results show that HiCT significantly outperforms existing baselines in PSNR, SSIM, and visual quality—even under sparse-view conditions. By bridging the gap between low-cost 2D PX and high-resolution 3D CBCT, HiCT enables accurate, low-radiation CBCT reconstruction, supporting intelligent clinical workflows in resource-limited settings and opening new possibilities for digital dental imaging.

\small
\bibliographystyle{IEEEbib}
\bibliography{strings,refs}

@article{px,
  title={Recent development in x-ray imaging technology: Future and challenges},
  author={Ou, Xiangyu and Chen, Xue and Xu, Xianning and Xie, Lili and Chen, Xiaofeng and Hong, Zhongzhu and Bai, Hua and Liu, Xiaowang and Chen, Qiushui and Li, Lin and others},
  journal={Research},
  year={2021},
  publisher={AAAS}
}

@inproceedings{Meshsnet,
  title={Meshsnet: Deep multi-scale mesh feature learning for end-to-end tooth labeling on 3d dental surfaces},
  author={Lian, Chunfeng and Wang, Li and Wu, Tai-Hsien and Liu, Mingxia and Dur{\'a}n, Francisca and Ko, Ching-Chang and Shen, Dinggang},
  booktitle={International Conference on Medical Image Computing and Computer-Assisted Intervention},
  pages={837--845},
  year={2019},
  organization={Springer}
}

@article{nerf,
  title={Nerf: Representing scenes as neural radiance fields for view synthesis},
  author={Mildenhall, Ben and Srinivasan, Pratul P and Tancik, Matthew and Barron, Jonathan T and Ramamoorthi, Ravi and Ng, Ren},
  journal={Communications of the ACM},
  volume={65},
  number={1},
  pages={99--106},
  year={2021},
  publisher={ACM New York, NY, USA}
}

@inproceedings{aenerf,
  title={AE-NeRF: Augmenting Event-Based Neural Radiance Fields for Non-ideal Conditions and Larger Scenes},
  author={Feng, Chaoran and Yu, Wangbo and Cheng, Xinhua and Tang, Zhenyu and Zhang, Junwu and Yuan, Li and Tian, Yonghong},
  booktitle={Proceedings of the AAAI Conference on Artificial Intelligence},
  volume={39},
  number={3},
  pages={2924--2932},
  year={2025}
}

@article{Nerf-det++,
  title={Nerf-det++: Incorporating semantic cues and perspective-aware depth supervision for indoor multi-view 3d detection},
  author={Huang, Chenxi and Hou, Yuenan and Ye, Weicai and Huang, Di and Huang, Xiaoshui and Lin, Binbin and Cai, Deng},
  journal={IEEE Transactions on Image Processing},
  year={2025},
  publisher={IEEE}
}

@inproceedings{zero123,
  title={Zero-1-to-3: Zero-shot one image to 3d object},
  author={Liu, Ruoshi and Wu, Rundi and Van Hoorick, Basile and Tokmakov, Pavel and Zakharov, Sergey and Vondrick, Carl},
  booktitle={Proceedings of the IEEE/CVF international conference on computer vision},
  pages={9298--9309},
  year={2023}
}

@article{ssim,
  title={Image quality assessment: from error visibility to structural similarity},
  author={Wang, Zhou and Bovik, Alan C and Sheikh, Hamid R and Simoncelli, Eero P},
  journal={IEEE transactions on image processing},
  volume={13},
  number={4},
  pages={600--612},
  year={2004},
  publisher={IEEE}
}

@article{Syncdreamer,
  title={Syncdreamer: Generating multiview-consistent images from a single-view image},
  author={Liu, Yuan and Lin, Cheng and Zeng, Zijiao and Long, Xiaoxiao and Liu, Lingjie and Komura, Taku and Wang, Wenping},
  journal={arXiv preprint arXiv:2309.03453},
  year={2023}
}

@inproceedings{yu2021pixelnerf,
  title={pixelnerf: Neural radiance fields from one or few images},
  author={Yu, Alex and Ye, Vickie and Tancik, Matthew and Kanazawa, Angjoo},
  booktitle={Proceedings of the IEEE/CVF conference on computer vision and pattern recognition},
  pages={4578--4587},
  year={2021}
}

@inproceedings{li2022infinitenature,
  title={Infinitenature-zero: Learning perpetual view generation of natural scenes from single images},
  author={Li, Zhengqi and Wang, Qianqian and Snavely, Noah and Kanazawa, Angjoo},
  booktitle={European conference on computer vision},
  pages={515--534},
  year={2022},
  organization={Springer}
}

@inproceedings{kulhanek2022viewformer,
  title={Viewformer: Nerf-free neural rendering from few images using transformers},
  author={Kulh{\'a}nek, Jon{\'a}{\v{s}} and Derner, Erik and Sattler, Torsten and Babu{\v{s}}ka, Robert},
  booktitle={European Conference on Computer Vision},
  pages={198--216},
  year={2022},
  organization={Springer}
}

@inproceedings{GeNVS,
  title={Generative novel view synthesis with 3d-aware diffusion models},
  author={Chan, Eric R and Nagano, Koki and Chan, Matthew A and Bergman, Alexander W and Park, Jeong Joon and Levy, Axel and Aittala, Miika and De Mello, Shalini and Karras, Tero and Wetzstein, Gordon},
  booktitle={Proceedings of the IEEE/CVF International Conference on Computer Vision},
  pages={4217--4229},
  year={2023}
}

@inproceedings{tseng,
  title={Consistent view synthesis with pose-guided diffusion models},
  author={Tseng, Hung-Yu and Li, Qinbo and Kim, Changil and Alsisan, Suhib and Huang, Jia-Bin and Kopf, Johannes},
  booktitle={Proceedings of the IEEE/CVF Conference on Computer Vision and Pattern Recognition},
  pages={16773--16783},
  year={2023}
}

@inproceedings{long2024wonder3d,
  title={Wonder3d: Single image to 3d using cross-domain diffusion},
  author={Long, Xiaoxiao and Guo, Yuan-Chen and Lin, Cheng and Liu, Yuan and Dou, Zhiyang and Liu, Lingjie and Ma, Yuexin and Zhang, Song-Hai and Habermann, Marc and Theobalt, Christian and others},
  booktitle={Proceedings of the IEEE/CVF conference on computer vision and pattern recognition},
  pages={9970--9980},
  year={2024}
}

@inproceedings{voleti2024sv3d,
  title={Sv3d: Novel multi-view synthesis and 3d generation from a single image using latent video diffusion},
  author={Voleti, Vikram and Yao, Chun-Han and Boss, Mark and Letts, Adam and Pankratz, David and Tochilkin, Dmitry and Laforte, Christian and Rombach, Robin and Jampani, Varun},
  booktitle={European Conference on Computer Vision},
  pages={439--457},
  year={2024},
  organization={Springer}
}

@inproceedings{lin2023learning,
  title={Learning deep intensity field for extremely sparse-view cbct reconstruction},
  author={Lin, Yiqun and Luo, Zhongjin and Zhao, Wei and Li, Xiaomeng},
  booktitle={International Conference on Medical Image Computing and Computer-Assisted Intervention},
  pages={13--23},
  year={2023},
  organization={Springer}
}

@inproceedings{chung2023solving,
  title={Solving 3d inverse problems using pre-trained 2d diffusion models},
  author={Chung, Hyungjin and Ryu, Dohoon and McCann, Michael T and Klasky, Marc L and Ye, Jong Chul},
  booktitle={Proceedings of the IEEE/CVF conference on computer vision and pattern recognition},
  pages={22542--22551},
  year={2023}
}

@inproceedings{ds-nerf,
  title={Depth-supervised nerf: Fewer views and faster training for free},
  author={Deng, Kangle and Liu, Andrew and Zhu, Jun-Yan and Ramanan, Deva},
  booktitle={Proceedings of the IEEE/CVF conference on computer vision and pattern recognition},
  pages={12882--12891},
  year={2022}
}

@inproceedings{shi2024zerorf,
  title={Zerorf: Fast sparse view 360deg reconstruction with zero pretraining},
  author={Shi, Ruoxi and Wei, Xinyue and Wang, Cheng and Su, Hao},
  booktitle={Proceedings of the IEEE/CVF Conference on Computer Vision and Pattern Recognition},
  pages={21114--21124},
  year={2024}
}

@inproceedings{huang2024neusurf,
  title={NeuSurf: On-surface priors for neural surface reconstruction from sparse input views},
  author={Huang, Han and Wu, Yulun and Zhou, Junsheng and Gao, Ge and Gu, Ming and Liu, Yu-Shen},
  booktitle={Proceedings of the AAAI conference on artificial intelligence},
  volume={38},
  number={3},
  pages={2312--2320},
  year={2024}
}

@article{zero123++,
  title={Zero123++: a single image to consistent multi-view diffusion base model},
  author={Shi, Ruoxi and Chen, Hansheng and Zhang, Zhuoyang and Liu, Minghua and Xu, Chao and Wei, Xinyue and Chen, Linghao and Zeng, Chong and Su, Hao},
  journal={arXiv preprint arXiv:2310.15110},
  year={2023}
}

@inproceedings{NAF,
  title={NAF: neural attenuation fields for sparse-view CBCT reconstruction},
  author={Zha, Ruyi and Zhang, Yanhao and Li, Hongdong},
  booktitle={International Conference on Medical Image Computing and Computer-Assisted Intervention},
  pages={442--452},
  year={2022},
  organization={Springer}
}

@article{FDK,
  title={Practical cone-beam algorithm},
  author={Feldkamp, Lee A and Davis, Lloyd C and Kress, James W},
  journal={Journal of the Optical Society of America A},
  volume={1},
  number={6},
  pages={612--619},
  year={1984},
  publisher={Optical Society of America}
}

@article{ASD,
  title={Image reconstruction in circular cone-beam computed tomography by constrained, total-variation minimization},
  author={Sidky, Emil Y and Pan, Xiaochuan},
  journal={Physics in Medicine \& Biology},
  volume={53},
  number={17},
  pages={4777},
  year={2008},
  publisher={IOP Publishing}
}

@article{liu2023deep,
  title={Deep learning-enabled 3D multimodal fusion of cone-beam CT and intraoral mesh scans for clinically applicable tooth-bone reconstruction},
  author={Liu, Jiaxiang and Hao, Jin and Lin, Hangzheng and Pan, Wei and Yang, Jianfei and Feng, Yang and Wang, Gaoang and Li, Jin and Jin, Zuolin and Zhao, Zhihe and others},
  journal={Patterns},
  volume={4},
  number={9},
  year={2023},
  publisher={Elsevier}
}

@inproceedings{liu2023toothsegnet,
  title={Toothsegnet: image degradation meets tooth segmentation in cbct images},
  author={Liu, Jiaxiang and Hu, Tianxiang and Feng, Yang and Ding, Wanghui and Liu, Zuozhu},
  booktitle={2023 IEEE 20th International Symposium on Biomedical Imaging (ISBI)},
  pages={1--5},
  year={2023},
  organization={IEEE}
}

@inproceedings{mw,
  title={PX2Tooth: Reconstructing the 3D Point Cloud Teeth from a Single Panoramic X-Ray},
  author={Ma, Wen and Wu, Huikai and Xiao, Zikai and Feng, Yang and Wu, Jian and Liu, Zuozhu},
  booktitle={International Conference on Medical Image Computing and Computer-Assisted Intervention},
  pages={411--421},
  year={2024},
  organization={Springer}
}

@article{covidx,
  author={Gunraj, Hayden and Sabri, Ali and Koff, David and Wong, Alexander},
  title={COVID-Net CT-2: Enhanced Deep Neural Networks for Detection of COVID-19 From Chest CT Images Through Bigger, More Diverse Learning},
  journal={Frontiers in Medicine},
  volume={8},
  pages={729287},
  year={2022},
  url={https://www.frontiersin.org/articles/10.3389/fmed.2021.729287},
  doi={10.3389/fmed.2021.729287},
  issn={2296-858X}
}

@article{TCIA,
  author = {Stanislav Nikolov and Sam Blackwell and Ruheena Mendes and Jeffrey De Fauw and Clemens Meyer and Cían Hughes and Harry Askham and Bernardino Romera-Paredes and Alan Karthikesalingam and Carlton Chu and Dawn Carnell and Cheng Boon and Derek D'Souza and Syed Ali Moinuddin and Kevin Sullivan and DeepMind Radiographer Consortium and Hugh Montgomery and Geraint Rees and Ricky Sharma and Mustafa Suleyman and Trevor Back and Joseph R. Ledsam and Olaf Ronneberger},
  title = {Deep learning to achieve clinically applicable segmentation of head and neck anatomy for radiotherapy},
  journal = {ArXiv e-prints},
  archivePrefix = "arXiv",
  eprint = {1809.04430},
  primaryClass = "cs.CV",
  year = {2018},
  url = {https://arxiv.org/abs/1809.04430}
}

\end{document}